\documentclass{article}

\usepackage{arxiv}

\usepackage[utf8]{inputenc} 
\usepackage[T1]{fontenc}    
\usepackage{hyperref}       
\usepackage{url}            
\usepackage{booktabs}       
\usepackage{amsfonts}       
\usepackage{nicefrac}       
\usepackage{microtype}      
\usepackage{cleveref}       
\usepackage{lipsum}         
\usepackage{graphicx}
\usepackage{natbib}
\usepackage{doi}

\title{RAN-JEPA: An Action-Conditioned World Model for Energy-efficient Mobile Network Control}

\newif\ifuniqueAffiliation
\uniqueAffiliationtrue

\ifuniqueAffiliation 
\author{ \href{https://orcid.org/0000-0001-6405-0873}{\includegraphics[scale=0.06]{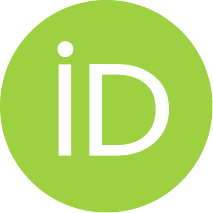}\hspace{1mm} Binghan Wu} \\
	AsiaInfo Technologies\\
	Beijing, China\\
	\texttt{wubh3@asiainfo.com} \\
	\And
	{Kai Zeng} \\
	AsiaInfo Technologies\\
	Beijing, China\\
	\texttt{zengkai@asiainfo.com}\\
	\And
	{Kai Yang} \\
	School of Computer Science and Technology\\
	Tongji University, China\\
	Shenzhen Loop Area Institute, China\\
	\texttt{kaiyang@tongji.edu.cn}\\
	\And
	{Ya-Qin Zhang} \\
	Institute for AI Industry Research (AIR)\\
	Tsinghua University\\
	Beijing, China\\
	\texttt{zhangyaqin@air.tsinghua.edu.cn}\\
	\And
	{Joseph Sifakis} \\
	Verimag\\
	Université Grenoble Alpes\\
	Grenoble, France\\
	\texttt{joseph.sifakis@univ-grenoble-alpes.fr}\\
	\And
	{Ye Ouyang} \\
	Institute for AI Industry Research (AIR)\\ Tsinghua University\\
	Beijing, China\\
	\texttt{ouyangye@air.tsinghua.edu.cn}\\
}
\else
\fi

\hypersetup{
pdftitle={A template for the arxiv style},
pdfsubject={q-bio.NC, q-bio.QM},
pdfauthor={David S.~Hippocampus, Elias D.~Striatum},
pdfkeywords={First keyword, Second keyword, More},
}

\begin{document}
\maketitle

\begin{abstract}
Operational mobile networks are rich in observations but poor in interventions: they continuously generate large volumes of passive telemetry, while controlled action--response trajectories remain comparatively scarce and costly to collect. This observation--intervention asymmetry is a central obstacle to building practical world models for RAN control. We address this problem with RAN-JEPA, an action-conditioned world model that separates representation learning from intervention learning. The model first learns a predictive latent representation from observation-only network data, adapts that representation to the target deployment domain, and then learns action-conditioned latent dynamics from controlled sleep/wake trajectories while keeping the encoder fixed. We evaluate RAN-JEPA using public operator performance data for pretraining and a real-equipment four-cell 5G testbed for domain adaptation, post-training, and control. Across four future 15-minute intervals, RAN-JEPA achieves lower action-conditioned prediction error than a GPT-2 world-model baseline and a mechanistic rule, while remaining close to a strong persistence reference. What-if evaluation further shows that the model produces distinct and coherent future trajectories under matched SLEEP and KEEP plans. In closed-loop control, RAN-JEPA consistently supports higher sleep ratios than the two baselines at comparable congestion cost. These results show that observation-first, action-conditioned learning can turn abundant passive telemetry and limited intervention data into a practical predictive model for energy-aware RAN control.
\end{abstract}


\section{Introduction}

Energy consumption has become an important operational concern for radio access networks (RANs) as mobile networks continue to expand in scale and capacity. One practical approach is to place selected cells or carriers into energy-saving states during periods of low demand and reactivate them when additional capacity is needed~\cite{niu2010cellzooming,salahdine2021sleep,3gpp28310r18}. Such actions reduce energy consumption but may also redistribute traffic, increase the load on neighboring cells, and affect user performance. Effective RAN energy saving therefore depends on anticipating the network response before an action is applied.

Conventional forecasting models estimate future traffic or performance from recent observations~\cite{zhang2018mobiletraffic}. Energy-saving control requires an additional capability because the controller may deliberately change the network configuration. For a given operating state, it must estimate how the network would evolve under different candidate actions, such as maintaining the current configuration, deactivating a cell, or reactivating previously suspended resources. This naturally leads to a what-if prediction problem.

A learned world model provides a useful abstraction for this purpose, building on earlier work on learned environment dynamics and latent-space planning~\cite{ha2018world,hafner2019planet}. It predicts future network behavior conditioned on the current state and a candidate action, while the final decision can be made by an external planner or controller. Recent work has demonstrated the feasibility of this approach for mobile-network energy saving~\cite{bouton2026mobile}. Compared with a policy that directly maps states to actions, a world model retains an explicit predictive representation of network dynamics and can be queried under different control objectives.

The main practical constraint is the availability of training data. Mobile operators continuously collect performance measurements, traffic statistics, resource-utilization counters, configuration states, and other operational telemetry. Most of these data describe the network under its existing operating policies. Controlled action-response trajectories are much less common because deliberate interventions require careful execution and monitoring and may affect capacity or user experience. We refer to this imbalance as the \emph{observation--intervention asymmetry} of telecom world-model learning.

This asymmetry suggests a staged learning process, analogous to action-free pretraining followed by action-conditioned world-model learning in visual control~\cite{seo2022actionfree}. Passive observations can first be used to learn a representation of the network and its regular temporal dynamics. Controlled intervention data can then be used to learn how specific actions modify those dynamics. The two datasets therefore serve different purposes rather than being treated as interchangeable training samples.

Joint-Embedding Predictive Architectures (JEPA) provide a suitable foundation for this setting. JEPA learns predictive relations in representation space instead of reconstructing all details of the original observation~\cite{lecun2022path,assran2023ijepa}. V-JEPA extended this idea to temporal observations~\cite{bardes2024vjepa}, while V-JEPA~2 showed that an action-free model can be pretrained from large-scale observations and later adapted into an action-conditioned world model using a smaller amount of interaction data~\cite{assran2025vjepa2}. This observation-first, action-later training principle is well aligned with the data available in operational communication networks.

We build on this principle and present \textbf{RAN-JEPA}, an action-conditioned world model for RAN energy saving. RAN-JEPA is first pretrained on passive network observations to learn latent RAN representations and temporal dynamics. It is subsequently post-trained on controlled action-response trajectories so that energy-saving actions explicitly condition future-state prediction. During inference, candidate actions can be evaluated through their predicted effects on network KPIs before being passed to an external energy-aware controller.

This article makes three contributions. First, it formulates RAN energy saving as an action-conditioned world-modeling problem under the observation--intervention asymmetry of telecom data. Second, it develops RAN-JEPA as an observation-pretrained, action-conditioned learning framework that separates the learning of network dynamics from the learning of intervention effects. Third, it evaluates this approach using real operator performance data for pretraining and controlled experiments on real network equipment for post-training and testing. The study focuses on RAN energy saving, while the resulting observations may also inform world-model learning for other network-control problems with similar data characteristics.


\section{RAN Energy Saving as a World-Modeling Problem}

RAN energy-saving decisions depend on both the current network condition and the expected consequences of a control action. This section defines the prediction problem considered in this work and positions it relative to forecasting, reinforcement learning, and existing action-conditioned world models.

\subsection{State, Action, and Future Outcome}

Let $\mathbf{x}_{t-L+1:t}$ denote a recent history of RAN observations. Depending on the deployment, these observations may contain traffic measurements, resource utilization, throughput, active-user statistics, cell operating states, and other performance indicators.

An energy-saving action $\mathbf{a}$ changes the network configuration. Typical actions include maintaining the current state, placing a cell or carrier into an energy-saving state, and reactivating resources that are currently sleeping. Network evolution is also affected by exogenous factors such as traffic demand, mobility, temporal patterns, and radio conditions. These factors are conceptually distinct from controller actions, but their future values are not assumed to be available at inference time. Their recent effects are instead reflected in the observed history.

The prediction problem is written as
\begin{equation}
\hat{\mathbf{x}}_{t+1:t+H}
=
F\!\left(
\mathbf{x}_{t-L+1:t},
\mathbf{a}_{t:t+H}
\right),
\label{eq:world_model_problem}
\end{equation}
where $F$ represents the learned network dynamics and $H$ is the prediction horizon. Given the same observed history, the model can therefore be queried with alternative candidate action plans to estimate how the network may evolve under each one.

This separation between observed state and planned action is important for control-oriented prediction. The action sequence represents an intervention selected by the controller, whereas traffic, mobility, and radio variation remain part of the operating environment. RAN-JEPA does not rely on oracle knowledge of their future realization; the resulting uncertainty is part of the prediction problem.

The learned dynamics model is also separated from the decision mechanism. RAN-JEPA predicts the network consequences of candidate actions. A planner, optimizer, or rule-based controller can then combine these predictions with an energy objective and performance constraints to select an action. The same world model can therefore support different control strategies without embedding a fixed decision policy in the predictive model itself.

\subsection{Existing Approaches and the Missing Piece}

Time-series forecasting has been widely used to predict traffic, load, and network performance~\cite{zhang2018mobiletraffic}. These models are effective when future evolution follows patterns represented in historical observations. Energy-saving control introduces a different setting because the controller deliberately changes the network configuration. Historical configuration variables may improve forecasting, but they mainly reflect the policies that generated the data and provide limited evidence about the consequences of alternative interventions.

Model-free reinforcement learning addresses the control problem by learning a policy or value function and has been extensively studied in communication networks~\cite{luong2019drlsurvey}. Its learned representation and behavior are typically optimized for a particular reward and policy. Changes in the control objective or operational constraints may therefore require policy retraining.

Action-conditioned world models instead represent how future network behavior depends on candidate actions, following the broader model-based control approach~\cite{ha2018world,hafner2019planet}. This makes them suitable for what-if evaluation and model-based control. Bouton \emph{et al.} demonstrated such a model for mobile-network energy saving~\cite{bouton2026mobile}. Related studies have also applied latent predictive learning to wireless channel dynamics and predictive remote control over capacity-limited networks~\cite{chaaya2025latentwireless,naoumi2026structured,girgis2026tsjepa}.

A practical difficulty is that action-conditioned trajectories are much less abundant than passive network telemetry. Operational networks continuously generate measurements of traffic, resource utilization, and service state, whereas controlled sleep/wake interventions are comparatively costly to collect. Passive observations still contain useful information about temporal regularities, dependencies among performance variables, and the natural evolution of network load, even when explicit intervention labels are absent.

RAN-JEPA is designed around this observation--intervention asymmetry. Observation-only data are first used to learn a predictive representation of RAN state and temporal evolution, including adaptation to the target deployment domain. Controlled action--response trajectories are introduced only afterward to learn how explicit energy-saving actions alter that evolution. The next section describes this staged training process.

\section{RAN-JEPA: From Observation to Action-Conditioned Dynamics}
\label{sec:method}

RAN-JEPA follows the JEPA principle of learning predictive structure in a latent space rather than reconstructing every network measurement~\cite{lecun2022path,assran2023ijepa,bardes2024vjepa}. Its main design choice is the training sequence, motivated by observation-first approaches to world-model learning~\cite{seo2022actionfree,assran2025vjepa2}. Observation-only data are used first to learn a transferable representation of RAN state and temporal evolution. The encoder can then be adapted to the target network using additional observation-only measurements from that domain. After this observation-only stage, the encoder is frozen and a separate action-conditioned rollout model is trained from controlled intervention trajectories. Only the encoder is carried into the action-conditioned stage; the rollout model and downstream prediction heads are trained from scratch.

\subsection{Observation-Only Predictive Learning}
\label{sec:method-pretraining}

The first stage uses network observation sequences without energy-saving action labels. A context encoder maps a recent window of RAN measurements to a latent representation of the current network state. A pretraining predictor estimates the representation of a future observation window, while a target encoder provides the corresponding training target and is updated as a slowly evolving counterpart of the context encoder.

Prediction is performed in representation space. The model is therefore not required to reconstruct every PM counter independently. Instead, the predictive objective encourages the encoder to preserve information that is useful for describing network evolution, including temporal traffic patterns, dependencies between traffic and resource utilization, and correlations among neighboring cells.

The same observation-only objective is also used for domain adaptation before action-conditioned training. This step exposes the encoder to the measurement distribution of the target network without requiring controlled sleep/wake trajectories. In our implementation, large-scale operator data are used for pretraining and observation-only measurements from the real-equipment testbed are then used for domain adaptation.

At the end of this stage, only the context encoder is retained. The pretraining predictor and target encoder are not used for downstream rollout or control. This separation keeps the roles of the two data regimes distinct: passive observations establish the representation, while intervention data are reserved for learning how explicit control actions modify future network behavior.

\subsection{Action-Conditioned Post-Training}
\label{sec:method-posttraining}

The second stage uses controlled action--response trajectories. The observation-trained encoder is frozen, and a new action-conditioned rollout predictor is initialized and trained together with the downstream heads. Given the recent observation history, the frozen encoder produces the initial latent state. The rollout predictor then propagates that state over the planned control intervals using the candidate energy-saving actions.

For rollout segment $q$, the transition is implemented as a residual update,
\begin{equation}
\hat{\mathbf{z}}^{(q+1)}
=
\hat{\mathbf{z}}^{(q)}
+
P_{\phi}\!\left(\hat{\mathbf{z}}^{(q)},\mathbf{a}^{(q)}\right),
\qquad
\hat{\mathbf{z}}^{(0)}=E_{\theta}(\mathbf{x}_{t-L+1:t}),
\label{eq:ran-jepa-rollout}
\end{equation}
where $E_{\theta}$ is the frozen encoder, $P_{\phi}$ is the action-conditioned rollout predictor, and $\mathbf{a}^{(q)}$ specifies the planned cell state for the corresponding future interval.

Only past observations and the future action plan are supplied to the model. Future traffic demand, mobility, and radio conditions are not provided as oracle inputs. Their effects are reflected through the observed history and through uncertainty in the predicted evolution. We nevertheless distinguish these exogenous factors conceptually from controller actions: a sleep or wake command is an intervention chosen by the controller, whereas traffic, mobility, and radio variation describe the environment in which that intervention takes effect.

This design focuses the controlled dataset on the part of the problem that passive telemetry cannot identify directly: the state-dependent effect of intervention. Deactivating the same cell can produce different load redistribution depending on the current UE population and neighboring-cell utilization. The action-conditioned rollout predictor learns these differences while operating on the representation established during observation-only learning.

\subsection{Prediction Heads and What-If Evaluation}
\label{sec:method-inference}

Each rolled latent state is mapped to two outputs. The numeric head predicts five controller-facing KPIs for each cell: UE count, downlink PRB utilization, uplink PRB utilization, downlink data volume, and uplink data volume. The direction head predicts whether each KPI increases or decreases relative to the same recent observation reference used to construct the numeric target.

The numeric prediction combines a mechanistic reference with a learned residual. Let $\tilde{\mathbf{y}}^{(q)}$ denote the reference produced from the recent measured state and the candidate action, and let $R_{\psi}$ be the learned numeric head. The final prediction is
\begin{equation}
\hat{\mathbf{y}}^{(q)}
=
\tilde{\mathbf{y}}^{(q)}
+
R_{\psi}\!\left(\hat{\mathbf{z}}^{(q)}\right).
\label{eq:ran-jepa-residual-head}
\end{equation}
The mechanistic component supplies a simple action-aware reference, while the learned head models the residual response that is not captured by that reference. In the implementation used in Section~\ref{sec:experiment}, both the numeric and direction targets are referenced to the most recent 15-minute mean.

At inference time, the required components are the frozen context encoder, the action-conditioned rollout predictor, and the two task heads. The target encoder and the observation-only pretraining predictor are used only during the first stage. Starting from the same observed history, the rollout model can be queried with different candidate action plans to obtain alternative future trajectories. This is the mechanism used for what-if comparison before an action is executed.

Energy consumption is not one of the five learned KPI outputs. Energy enters the control loop through the executed cell configuration and the external energy-accounting rule used by the controller. RAN-JEPA provides the predicted network consequences of each candidate action; the controller combines those predictions with its energy objective and QoS constraints when selecting an action.

Section~\ref{sec:experiment} describes the cross-domain observation-only training, controlled action collection, and evaluation protocol used for prediction and closed-loop control.


\providecommand{\experimentTODO}[1]{\textbf{[TODO: #1]}}
\providecommand{\experimentFigure}[3]{%
  \begin{figure}[htbp]
    \centering
    \IfFileExists{#1}{%
      \includegraphics[width=0.7\linewidth]{#1}%
    }{%
      \fbox{\parbox[c][35mm][c]{0.7\linewidth}{%
        \centering Figure placeholder\\[6pt]\texttt{\detokenize{#1}}}}%
    }
    \caption{#2}
    \label{#3}
  \end{figure}%
}

\section{Experiment}
\label{sec:experiment}

We evaluate RAN-JEPA in three steps. First, we measure action-conditioned prediction accuracy over a 60-minute horizon. Second, we use a calibrated digital twin to examine a representative sleep episode under both the executed SLEEP action and a matched KEEP alternative. Third, we place the trained model in the same closed-loop controller used by the baselines and measure the resulting energy--QoS trade-off. The first two learning stages use observation-only data, while action-conditioned post-training and control use trajectories collected from the controlled testbed.

\subsection{Data Sources and Testbed}
\label{sec:experiment-data}

\paragraph{Observation-only learning.}
The encoder is first pretrained on the CETIN performance-management dataset~\cite{lehoczky2026performance}. We use the complete Dataset\_03, comprising 30,213,561 PM records from 264 base stations over 84 days at 15-minute intervals. These records contain no explicit sleep-action labels in our training pipeline and are used with the observation-only predictive objective described in Section~\ref{sec:method-pretraining}.

The encoder uses the same 22-field input interface throughout observation-only learning and downstream training. Nineteen PM fields are shared by CETIN and the testbed. For CETIN, the three testbed-only fields are assigned the designated no-action value, so the encoder input dimensionality remains unchanged across data sources. Both sources use 12 consecutive observation windows per cell. Because their native sampling intervals are retained, the CETIN context spans 3~hours whereas the testbed context spans 1~hour.

After CETIN pretraining, the encoder is further adapted using 180 days of observation-only measurements collected on the real-equipment testbed. The same predictive learning objective is retained, but the data now come from the target deployment domain. This step is used for domain adaptation before any controlled sleep/wake trajectory is introduced.

\paragraph{Real-equipment testbed.}
The testbed in Fig.~\ref{fig:experiment-testbed} comprises a 5G core, a 5G NR base station, a TM500 MK4.4 multi-UE emulator, and a control workstation. The radio configuration uses n78 at 3.5~GHz, TDD, 100~MHz bandwidth, and 30~kHz subcarrier spacing. The experiments use four cells and between 0 and 500 emulated UEs. A programmable attenuation matrix with a range of 0--90~dB in 0.5~dB steps controls the RF paths. The attenuated cell signals are combined inside a shielded enclosure and fed to the TM500, allowing repeatable traffic and path-loss conditions while the base station runs its actual protocol stack.

\experimentFigure{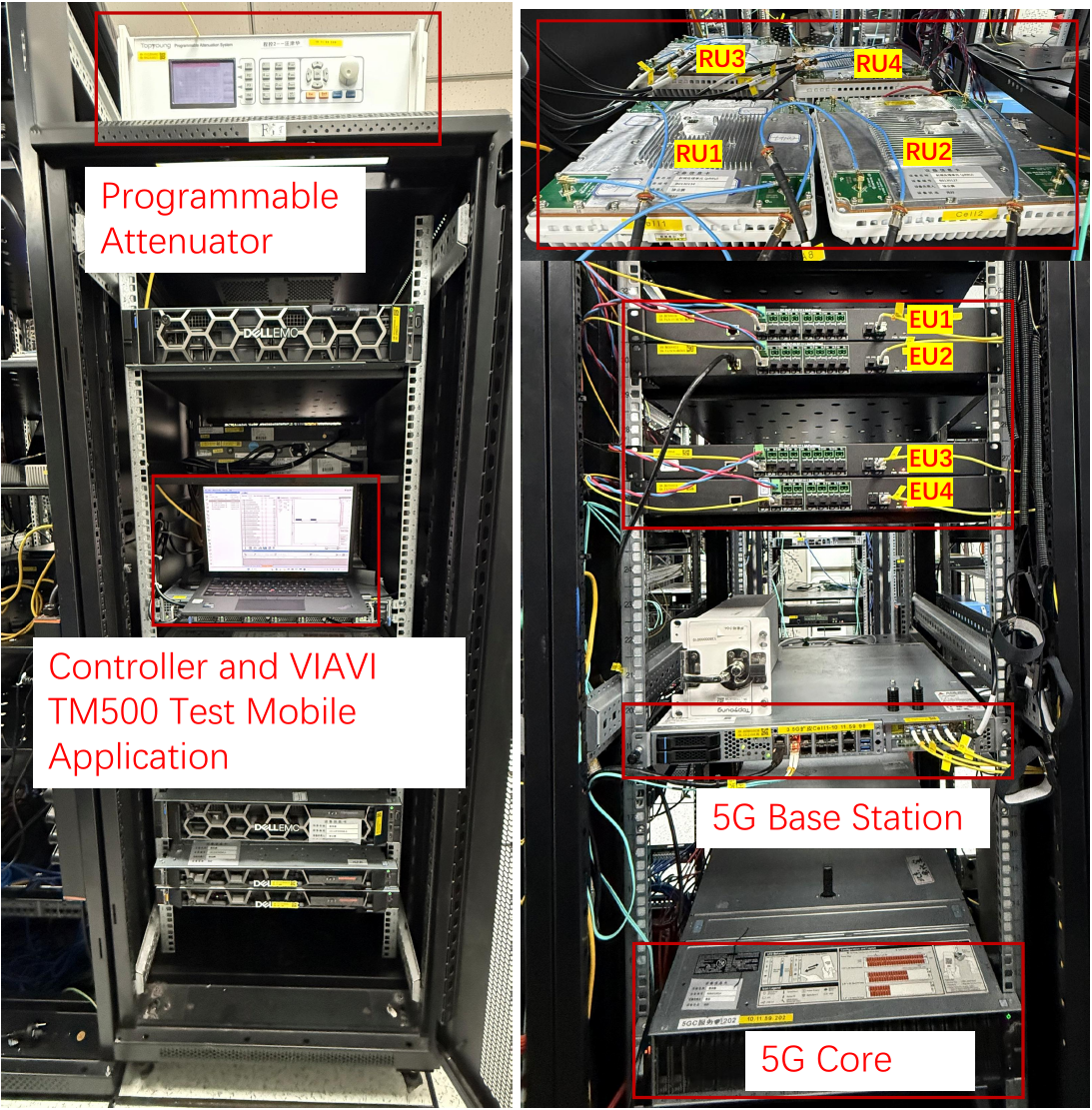}{%
Real-equipment testbed used to collect observation and action--response trajectories. Programmable attenuation controls the RF paths between the four cells and the TM500 UE emulator.}{fig:experiment-testbed}

Testbed KPIs are recorded every five minutes. The 22 downstream fields include the 19 fields shared with CETIN---covering user counts, PRB usage and availability, uplink and downlink traffic, and transmission-block counts grouped by size---plus three testbed-only fields. The five prediction targets are UE count, downlink PRB utilization, uplink PRB utilization, downlink data volume, and uplink data volume.

In addition to the 180-day observation-only collection used for domain adaptation, the testbed provides 90 days of controlled action--response trajectories. These trajectories are used for action-conditioned post-training and for the main prediction and control evaluation.

\paragraph{Sleep operation.}
Sleep is implemented as a protocol-level energy-saving state rather than a complete RF shutdown. The SSB remains available, while UEs served by a cell selected for sleep are unloaded to active neighboring cells through the normal handover procedure. One cell is kept active as the coverage anchor and the other three are eligible for sleep. Wake-up restores a sleeping cell to the active configuration before it is used again for traffic service.

\subsection{Training and Evaluation Protocol}
\label{sec:experiment-protocol}

After CETIN pretraining and testbed-domain adaptation, only the encoder weights are retained and the encoder is frozen. The action-conditioned latent rollout predictor, numeric head, and direction head are initialized from scratch and trained on the measured action--response trajectories. The GPT-2-style world-model baseline uses the sequence-modeling architecture introduced by Radford et al.~\cite{radford2019gpt2}, also used as a backbone in the mobile-network world model of Bouton et al.~\cite{bouton2026mobile}. Our baseline is trained from scratch throughout. No model weights are updated during open-loop evaluation or closed-loop control.

The action--response data are split by complete experimental scenario into training, validation, and test sets using a 70/15/15 ratio. All windows from the same scenario remain in the same partition, and input--target windows are never allowed to cross scenario boundaries. Open-loop learned-model results are reported over 15 runs spanning the prescribed data-split and random-seed variations. Closed-loop results use nine matched runs per controller setting. For the deterministic mechanistic rule, variation arises from the evaluated data partitions; for learned models, it reflects both partition and random-seed variation.

Each prediction uses twelve five-minute observation windows from four cells. The candidate action plan specifies the sleep state of each cell over four future 15-minute segments, yielding a 60-minute rollout. Only the past observation history and the planned actions are supplied to the model; future traffic and future measurements are not available at inference time. Numeric targets are expressed relative to the most recent 15-minute mean and evaluated in the original KPI units.

\experimentFigure{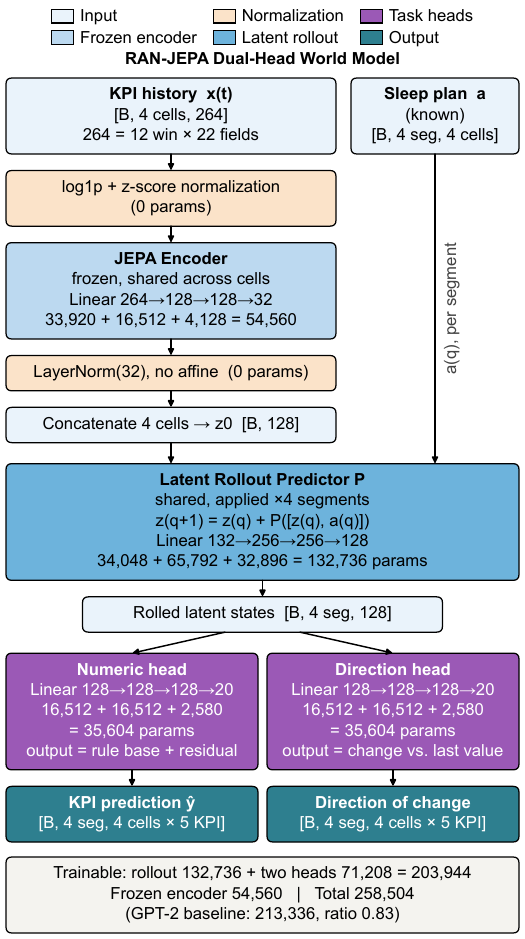}{%
RAN-JEPA implementation used in the experiments. A frozen observation-trained encoder maps the 12-window KPI history to a latent state. A newly trained action-conditioned rollout predictor propagates that state across four future segments, and numeric and direction heads produce the five downstream KPIs. The numeric head predicts a residual on top of the mechanistic reference.}{fig:ran-jepa-architecture}

We compare RAN-JEPA with two benchmarks: a mechanistic rule and a GPT-2-based world model. The persistence reference used to normalize MAE repeats the most recent 15-minute mean and is not treated as a trained benchmark. The mechanistic rule uses the previous-day reference together with the planned sleep action; consequently, the comparison in Fig.~\ref{fig:experiment-prediction} is restricted to samples for which that reference is available. GPT-2 receives the same observation history and window-level cell-state information, while RAN-JEPA receives the observation history and interval-level action plan. All methods are evaluated on the same eligible sleep-action samples.

GPT-2 has 213,336 parameters. RAN-JEPA has 258,504 parameters in total, of which 203,944 are trainable during action-conditioned post-training and 54,560 belong to the frozen encoder. For both learned methods, we test learning rates of $10^{-3}$ and $3\times10^{-4}$ and select the rate on the validation set.

\subsection{Prediction Across Control Horizons}
\label{sec:experiment-prediction}

Figure~\ref{fig:experiment-prediction} reports action-conditioned prediction error on neighboring cells after a sleep action. The evaluation is limited to samples with an available previous-day reference because that reference is required by the mechanistic rule. MAE is normalized by the persistence error at the same target and horizon, so a relative MAE below one indicates lower error than persistence.

\experimentFigure{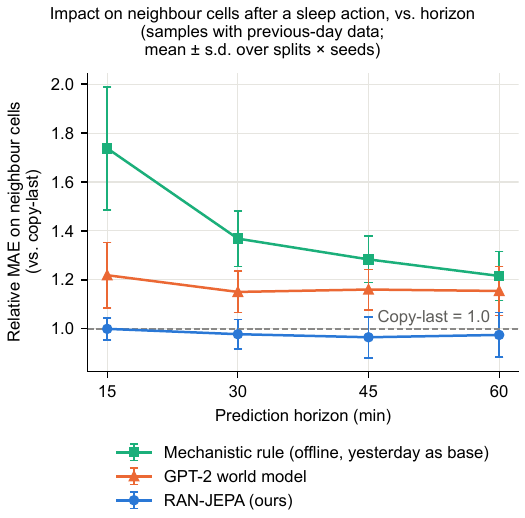}{%
Action-conditioned prediction error on neighboring cells across four future 15-minute intervals. Relative MAE is normalized by persistence, for which one denotes equal error. The evaluation includes only samples with an available previous-day reference. Error bars show one standard deviation across repeated evaluations.}{fig:experiment-prediction}

Across the four horizons, RAN-JEPA remains close to the persistence reference, with relative MAE of approximately 0.97--1.00. GPT-2 is higher at approximately 1.15--1.22, while the mechanistic rule ranges from approximately 1.22 to 1.74. RAN-JEPA therefore yields lower average error than the two action-conditioned baselines in this evaluation, while its margin relative to persistence remains small. The plotted variation does not support a claim of statistical significance against persistence.

The relative MAE decreases with horizon for some methods, but this does not imply that their absolute error decreases at longer horizons. Persistence error can itself grow with the prediction horizon. The result should therefore be read as a change in error relative to persistence rather than as evidence that longer-horizon prediction is intrinsically easier.

\subsection{What-If Analysis of a Sleep Intervention}
\label{sec:experiment-whatif}

Figure~\ref{fig:experiment-whatif} provides a direct what-if view of one sleep episode using the calibrated digital twin. The representative case is selected as the episode with median factual prediction error among 734 eligible episodes, rather than as a best-case example. The future trajectory is summarized in 15-minute intervals. Two quantities are shown: UE count on the action cell and downlink PRB utilization on a serving cell that receives redistributed load.

Starting from the same pre-action history, the model is queried with the executed SLEEP plan and with a KEEP plan. The calibrated twin provides a reference trajectory for both plans, allowing the factual and counterfactual branches to be inspected under matched initial conditions. Traces are reported as percentage change relative to the value at action onset.

\experimentFigure{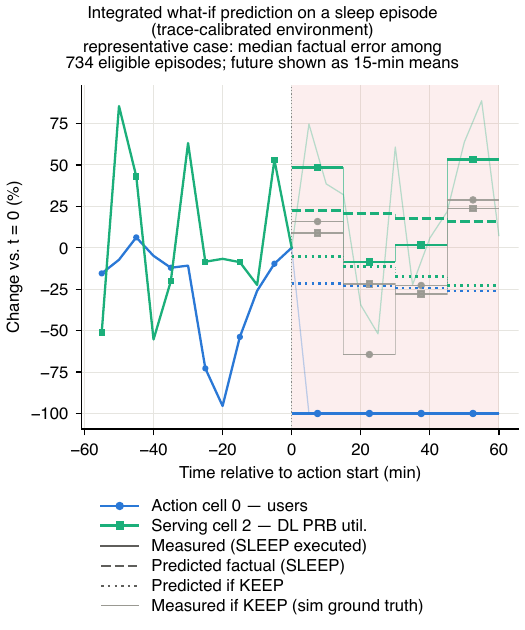}{%
Integrated what-if prediction for a representative sleep episode in the calibrated digital twin. The case has median factual error among 734 eligible episodes. The pre-action history is followed by four future 15-minute intervals. Solid trajectories show the reference evolution, dashed trajectories the factual prediction under SLEEP, and dotted trajectories the prediction under KEEP; the calibrated twin also provides a KEEP reference under the same initial condition.}{fig:experiment-whatif}

The action-cell behavior is captured cleanly in the factual branch: after SLEEP is applied, the UE count falls to $-100\%$ relative to the action-onset value and remains there across the prediction horizon, consistent with the cell no longer serving users. On the neighboring serving cell, the model predicts an immediate positive shift in downlink PRB utilization under SLEEP, reflecting traffic redistribution away from the action cell. The prediction captures the direction and sustained component of the load transfer, while the reference trajectory exhibits substantially larger short-term variation than the 15-minute predicted means.

The KEEP branch is materially different from the SLEEP branch for both quantities. In particular, the model does not impose the complete UE loss associated with putting the action cell to sleep, and it predicts a lower load on the neighboring serving cell. Because the calibrated digital twin supplies a matched KEEP reference, this panel tests not only factual prediction but also whether the learned dynamics respond coherently when the action is changed while the observed history is held fixed.

\subsection{Closed-Loop Energy--QoS Trade-off}
\label{sec:experiment-control}

The final experiment evaluates whether the prediction differences observed above translate into different closed-loop operating points. The mechanistic, GPT-2, and RAN-JEPA controllers use the same candidate-action generator and downstream decision logic; only the predictive model used to evaluate candidate sleep plans is changed. At each decision epoch, candidate plans are screened against the same QoS criterion, and the controller selects among the feasible plans according to the same energy-saving objective. The models are loaded once before each run and are not updated online.

Control runs are matched by traffic demand and attenuation trajectory so that the three controllers face the same exogenous conditions. Nine matched runs are used for each reported controller setting. We characterize the operating point using two quantities measured under the executed actions: the sleep ratio, which serves as the operational proxy for energy-saving aggressiveness, and the additional congestion time caused by sleeping, reported in percentage points as the QoS cost. A larger sleep ratio is desirable for a fixed congestion increase, whereas a smaller congestion increase is desirable for a fixed sleep ratio.

\experimentFigure{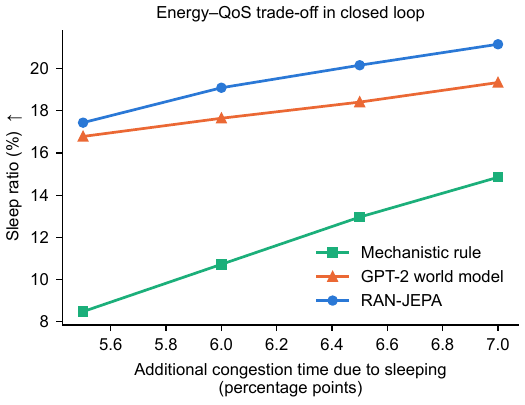}{%
Closed-loop energy--QoS trade-off on the real-equipment testbed. The vertical axis reports sleep ratio as the operational proxy for energy-saving aggressiveness, and the horizontal axis reports the additional congestion time caused by sleeping. Operating points correspond to matched controller settings under the same traffic and attenuation conditions.}{fig:experiment-control}

Figure~\ref{fig:experiment-control} shows a consistent separation among the three controllers across the evaluated QoS range. At an additional congestion time of about 5.5 percentage points, RAN-JEPA reaches a sleep ratio of approximately 17.4\%, compared with 16.8\% for GPT-2 and 8.5\% for the mechanistic rule. As the allowed congestion increase rises, all three controllers sleep more aggressively, but the ordering remains unchanged. At about 7.0 percentage points of additional congestion, the sleep ratios are approximately 21.1\% for RAN-JEPA, 19.3\% for GPT-2, and 14.9\% for the mechanistic rule.

The result therefore indicates that the predictive model affects the attainable closed-loop operating point rather than only the open-loop forecasting error. Across the four reported settings, RAN-JEPA supports a higher sleep ratio than GPT-2 at comparable congestion cost, with an advantage that grows from roughly 0.7 to 1.8 percentage points over the plotted range. The margin over the mechanistic controller is larger, ranging from roughly 6 to 9 percentage points of sleep ratio. These differences are descriptive of the evaluated matched runs; the figure does not by itself establish statistical significance.

Taken together with Figs.~\ref{fig:experiment-prediction} and~\ref{fig:experiment-whatif}, the closed-loop result completes the evaluation chain. RAN-JEPA first provides lower action-conditioned prediction error than the two action-aware baselines, then produces distinct and coherent trajectories when the candidate action is changed, and finally converts those predictions into a more favorable sleep--congestion trade-off in the controlled testbed. The measurements characterize four-cell control under repeatable laboratory conditions and should not be interpreted as network-wide energy savings or as performance guarantees for all commercial deployments.

\section{Conclusion}

This work studied RAN energy saving as an action-conditioned world-modeling problem under the observation--intervention asymmetry of telecom data. Rather than training directly on the limited set of controlled interventions, RAN-JEPA first learns a predictive representation from observation-only measurements, adapts that representation to the target deployment domain, and then learns action-dependent latent transitions from controlled sleep/wake trajectories. This separation allows passive telemetry to establish the network representation while reserving intervention data for learning the effects of control actions.

The evaluation covered prediction, what-if analysis, and closed-loop control. RAN-JEPA achieved lower action-conditioned prediction error than the GPT-2 and mechanistic baselines across the evaluated horizons, while remaining close to a strong persistence reference. In the calibrated digital-twin analysis, changing the candidate action from SLEEP to KEEP produced distinct future trajectories under the same observed history, providing evidence that the learned dynamics respond to the intervention rather than only extrapolating recent measurements. In closed-loop experiments on the real-equipment testbed, these prediction differences translated into a more favorable sleep--congestion trade-off, with RAN-JEPA supporting higher sleep ratios than the baselines at comparable congestion cost.

The present experiments are limited to a four-cell controlled testbed and use sleep ratio as an operational proxy for energy-saving aggressiveness; they should therefore not be interpreted as network-wide energy savings. A natural next step is to evaluate the same observation-first, action-conditioned training principle under larger topologies, broader operating conditions, and direct energy measurements. More generally, the results suggest that separating representation learning from intervention learning is a useful design pattern for telecom control problems in which observational data are plentiful but controlled actions are costly to collect.

\bibliographystyle{unsrtnat}
\bibliography{references}

\end{document}